\documentclass{article}

\usepackage{microtype}
\usepackage{graphicx}
\usepackage{subfigure}
\usepackage{booktabs} 

\usepackage{hyperref}

\usepackage[arxiv]{mlsys2020}

\usepackage{listings}
\usepackage{multirow}
\usepackage{url}

\mlsystitlerunning{VEGA: Towards an End-to-End Configurable AutoML Pipeline}

\begin{document}

\twocolumn[
\mlsystitle{VEGA: Towards an End-to-End Configurable AutoML Pipeline}



\mlsyssetsymbol{equal}{*}

\begin{mlsysauthorlist}
\mlsysauthor{Bochao Wang}{noah}
\mlsysauthor{Hang Xu}{noah}
\mlsysauthor{Jiajin Zhang}{noah}
\mlsysauthor{Chen Chen}{noah}
\mlsysauthor{Xiaozhi Fang}{noah}
\mlsysauthor{Yixing Xu}{noah}
\mlsysauthor{Ning Kang}{noah}
\mlsysauthor{Lanqing Hong}{noah}
\mlsysauthor{Chenhan Jiang}{noah}
\mlsysauthor{Xinyue Cai}{noah}
\mlsysauthor{Jiawei Li}{noah}
\mlsysauthor{Fengwei Zhou}{noah}
\mlsysauthor{Yong Li}{noah}
\mlsysauthor{Zhicheng Liu}{noah}
\mlsysauthor{Xinghao Chen}{noah}
\mlsysauthor{Kai Han}{noah}
\mlsysauthor{Han Shu}{noah}
\mlsysauthor{Dehua Song}{noah}
\mlsysauthor{Yunhe Wang}{noah}
\mlsysauthor{Wei Zhang}{noah}
\mlsysauthor{Chunjing Xu}{noah}
\mlsysauthor{Zhenguo Li}{noah}
\mlsysauthor{Wenzhi Liu}{noah}
\mlsysauthor{Tong Zhang}{hku}
\end{mlsysauthorlist}

\mlsysaffiliation{noah}{Huawei Noah's Ark Lab, China.}
\mlsysaffiliation{hku}{Hong Kong University of Science and Technology, China}

\mlsyscorrespondingauthor{Wei Zhang}{wz.zhang@huawei.com}

\mlsyskeywords{Machine Learning, MLSys}

\vskip 0.3in

\begin{abstract}
Automated Machine Learning (AutoML) is an important industrial solution for automatic discovery and deployment of the machine learning models. However, designing an integrated AutoML system faces four great challenges of configurability, scalability, integrability, and platform diversity. In this work, we present VEGA, an efficient and comprehensive AutoML framework that is compatible and optimized for multiple hardware platforms. a) The VEGA pipeline integrates various modules of AutoML, including Neural Architecture Search (NAS), Hyperparameter Optimization (HPO), Auto Data Augmentation, Model Compression, and Fully Train. b) To support a variety of search algorithms and tasks, we design a novel fine-grained search space and its description language to enable easy adaptation to different search algorithms and tasks. c) We abstract the common components of deep learning frameworks into a unified interface. VEGA can be executed with multiple back-ends and hardwares. Extensive benchmark experiments on multiple tasks demonstrate that VEGA can improve the existing AutoML algorithms and discover new high-performance models against SOTA methods, e.g. the searched DNet model zoo for Ascend 10x faster than EfficientNet-B5 and 9.2x faster than RegNetX-32GF on ImageNet. 
VEGA is open-sourced at \url{https://github.com/huawei-noah/vega}.
\end{abstract}
]



\printAffiliationsAndNotice{}  

\section{Introduction}
\label{sec:intro}
Automated Machine Learning (AutoML) \citep{he2019automl} is an integration
of machine learning algorithms that enables developers and users with
limited machine learning expertise to train and deploy high-quality
models for their own tasks and needs. It is an important industrial
solution for the fast and easy deployment of machine learning algorithms.
Nowadays,
deep learning has become the de facto method for artificial intelligence
in various tasks, such as computer vision \citep{HeZRS16,ren2015faster},
natural language processing (NLP) \citep{yin2017comparative,devlin2018bert}, autonomous driving \citep{al2017deep,kuutti2020survey}
and recommender systems \citep{guo2017deepfm,liu2020autogroup}.
New modules and architectures are required for deep learning tasks.
Neural architecture search (NAS) \citep{tan2019efficientnet,jiang2020sp,shi2020bonas},
hyperparameter optimization (HPO) \citep{falkner2018bohb} and data 
augmentation \citep{cubuk2018autoaugment} have made great progress in many tasks.
However, most efforts focus on only one or two modules of AutoML and thus are suboptimal.
A modern and integrated AutoML system is essential.

The major challenges for an efficient AutoML system can be categorized
into four aspects: integrability, configurability, scalability, and
platform diversity. Our VEGA framework provides our solutions for each
of the challenges:

\textbf{(a) Integrability.} A complete AutoML system should integrate
necessary modules and functions such as NAS, HPO,  data augmentation, and model compression \citep{he2018amc}.
Besides, new modules or operators \cite{tang2020beyond} need to be supported for more choices.
Note that NAS and data augmentation are cutting-edge deep learning techniques
and proved to be effective in applications \citep{xie2020self,YaoLewei19,xu2019auto,xu2020curvelane}.
VEGA integrates all those modules into one pipeline.
Each module can be implemented as a pipe step. 
VEGA can sequentially execute each step and output the final model as expected.

\textbf{(b) Configurability.} Each module in an AutoML algorithm usually has multiple choices.
It is tedious to hard code all components.
VEGA makes all components configurable flexibly by a hierarchical structure, where
developers can simply configure all information in a YML file.
It is easy to customize architectures, hyperparameters, and training processes in VEGA. 
For these components, VEGA provides a unified search space.
Actually, most search spaces are highly correlated with their own search algorithms, and these are not general and flexible enough to adapt to new tasks and applications.
For example, some weight-sharing
NAS methods such as DARTS \citep{liu2018darts} require a super-net
with DAG-based cell level search space, while some sample-based NAS
algorithms \citep{real2019regularized,YaoLewei19}
require an optimization of the network's width and depth.
 Obviously, those search spaces are separated and cannot be optimized simultaneously.
How to design a general and flexible search space that can be applied
to multiple search algorithms remains a challenging problem. 
To this end,  VEGA provides a fine-grained search space description language that can describe 
the hyperparameters of each module in networks from micro-level operators to marco-level architectures.
With this language, the defined search space can be independent of the used search algorithm.

\textbf{(c) Scalability.} 
It is known that NAS and HPO are compute-intensive.
AutoML requires huge resources when undertaking search-based
algorithms such as BOHB \citep{falkner2018bohb} and EA-based NAS
\citep{cai2019once}. 
VEGA abstracts the training process of models as a trainer and designs a
distributed search system that can easily manage the dispatch of tasks 
and the gather of evaluated results.

\textbf{(d) Platform Diversity. } Many deep learning models may be trained or
searched with GPUs and deployed on specific hardware, such as mobile devices.
VEGA supports multiple hardware platforms for model training or searching, 
such as CPU, GPU, Ascend \citep{liao2019davinci} and multiple deep learning frameworks as back-ends,
including Pytorch \citep{paszke2019pytorch}, Tensorflow \citep{abadi2016tensorflow} 
and MindSpore \footnote{\url{https://github.com/mindspore-ai/mindspore}}.
Besides, VEGA provides model conversion tools to assist the trained model to be
deployed on more devices by ONNX \citep{bai2019onnx}.
It is easy to collect the feedback of a model deployed on target platforms.

To deal with these challenges, we propose VEGA, an efficient and comprehensive
AutoML framework tailored to multiple back-ends and hardware platforms. 
Our contributions can be summarized as follows:
\begin{itemize}
\item The VEGA pipeline integrates multiple modules of AutoML,
i.e. neural architecture search (NAS), hyperparameter optimization (HPO), 
data augmentation and model compression.

\item To fit in various kinds of search algorithms and tasks,
we design a novel fine-grained search space and its description language
to enable easy adaptation to different scenarios. 

\item We introduce unified interfaces for common components of deep learning frameworks.
VEGA supports multiple frameworks and hardware platforms.

\item Extensive benchmark experiments in various tasks demonstrate that the developed VEGA pipeline can
improve the existing AutoML algorithms and generate high-performance
models against SOTA methods, e.g. the searched DNet model zoo for
Ascent is $10\times$ faster than EfficientNet-B5 \citep{tan2019efficientnet}
and $9.2\times$ faster than RegNetX-32GF \citep{radosavovic2020designing}
on ImageNet \citep{deng2009imagenet}. 
\end{itemize}


\section{Related Work}
\begin{table*}[h]
\vspace{-2mm}
\caption{Comparison between open sourced AutoML Pipelines. Our VEGA has more
flexible search space, more functions and tasks}
\label{tab:compare_other_pipeline}
\begin{centering}
\renewcommand\arraystretch{0.8}\tabcolsep 0.02in%
\begin{tabular}{ p{2cm} | p{4cm} | p{4cm} | p{6cm} }
\hline 
\textbf{\tiny{}AutoML Pipeline} & \textbf{\tiny{}NNI} & \textbf{\tiny{}AutoGluon} & {\tiny{}VEGA}\tabularnewline
\hline 
\textbf{\tiny{}Support Functions} & {\tiny{}5} & {\tiny{}3} & {\tiny{}7}\tabularnewline
\hline 
\textbf{\tiny{}NAS Search Space} & {\tiny{}Support DARTS cell and ENAS search space. Require users to
specify search space and the algorthim codes.} & {\tiny{}Support limited search space definition in GluonCV and GluonNLP.} & {\tiny{}Decoupled search space. Support commonly used network search
space including ResNet, DARTS, ESRN, Faster-RCNN and DNET. Different
search algorithms can be used for the same search space.}\tabularnewline
\hline 
\textbf{\tiny{}Number of Tasks} & {\tiny{}2} & {\tiny{}4} & {\tiny{}7}\tabularnewline
\hline 
\textbf{\tiny{}Tasks} & {\tiny{}Image Classification, Model Compression} & {\tiny{}Tabular Prediction, Image Classification, Object Detection,
Text Prediction} & {\tiny{}Image Classification, Detection, Lane Detection, Model Compression,
Image SR, Semantic segmentation, Click-Through Rate Prediction}\tabularnewline
\hline 
\end{tabular}
\par\end{centering}
\vspace{-2mm}
\end{table*}

AutoML has made great achievements in many modules, including neural architecture search (NAS), hyperparameters optimization (HPO), data augmentation, and model compression.
There are many surveys \cite{he2019automl,zoller2019benchmark,elsken2019neural,cheng2017survey,shorten2019survey} that have summarized classical algorithms on these topics.
Most of these algorithms focus on one or two modules and design their own search space.
These search spaces are strongly coupled with corresponding search algorithms.
It lacks a universal framework and a unified search space for various AutoML algorithms.

DeepArchitect \cite{negrinho2017deeparchitect} attempts to unify the representation of search spaces over architectures and their hyperparameters and then proposes an extensible and modular language.
However, it requires developers to re-implement the graph structure of architecture with the proposed language.
This language focuses on training hyperparameters and training hyperparameters.
Their available values are explicitly expressed in language.
It cannot be applied to complex scenarios, for example, weight-sharing algorithms.
NNI (Neural Network Intelligence) \footnote{\url{https://github.com/microsoft/nni}} is a more powerful toolkit and it can be used for NAS, HPO, and model compression.
The search space is statically defined with the name and choices of a variable.
It prefers to one-shot NAS algorithms, such as ENAS \cite{pham2018efficient} and DARTS \cite{liu2018darts}.
To define the search space of a customized network, developers need to define extra specific codes. Table \ref{tab:compare_other_pipeline}  shows some comparison between several well-known open-sourced AutoML Pipelines.
AutoGluon \cite{agtabular} is another AutoML toolkit and can be used for tabular prediction, image classification, object detection, and text prediction.
Similar to NNI, it's better at the search space of one-shot NAS.
It's hard to extend to the case with dynamic variables in NAS.

VEGA provides a fine-grained search space description language that can describe the hyperparameters of each module in networks from micro-level operators to macro-level architectures.
The search space of VEGA can be defined and adjusted through a configuration file and different search algorithms can be used in the same search space.
Besides, VEGA has a great range of application scenarios, including image
classification, objection/pedestrian detection, semantic segmentation, image
super-resolution, and click-through rate prediction, as listed in Section
\ref{sec:pipeline}.

\section{Architecture}
\label{sec:arch}
\subsection{Overview}
\label{sec:arch_overview}
\begin{figure}
   \vspace{-2mm}
  \centering
  \includegraphics[scale=0.28]{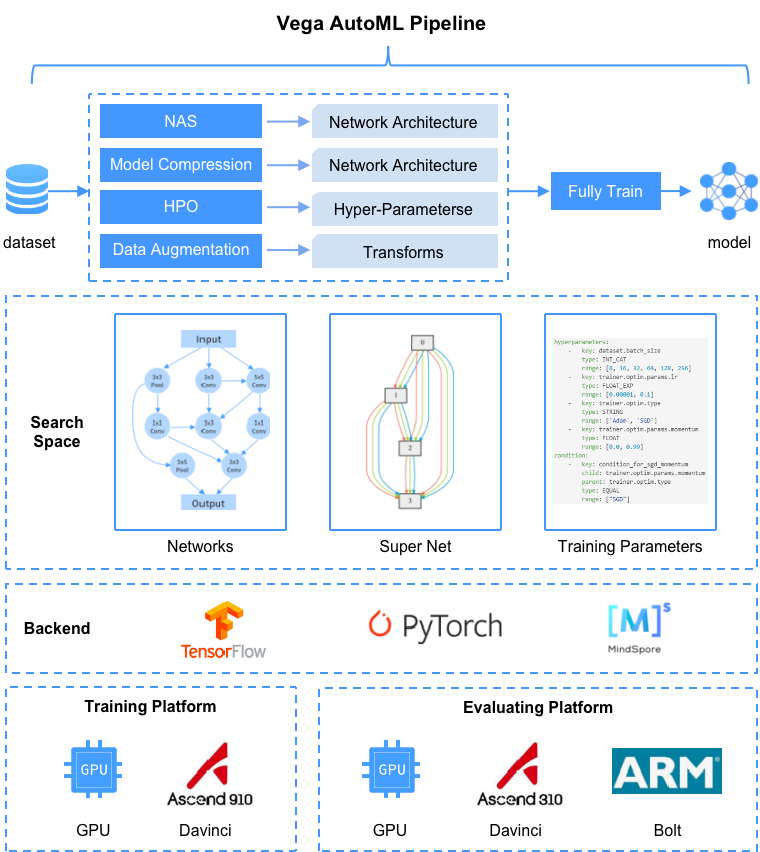}
    \vspace{-3mm}
  \caption{Overview of Our VEGA AutoML pipeline. We support various AutoML features, back-ends and hardware platforms. }
  \label{fig:pipeline}
   \vspace{-2mm}
\end{figure}

VEGA is an efficient and comprehensive AutoML system.
As illustrated in Figure \ref{fig:pipeline}, it integrates multiple modules of AutoML
into a unified framework, such as NAS, HPO, data augmentation, and model compression.
Fully train is a special step of the pipeline to get the final model or implement new modules for architectures.
VEGA treats these algorithms as different steps in the pipeline of building a deep
learning system.
The key to a good AutoML algorithm is how to define an efficient search space.
VEGA provides a unified and configurable search space prototype.
Hyperparameters for HPO and transformers for data augmentation can be easily
represented with their names and the range of values.
More details are discussed in Section \ref{sec:coarse_search_space}.
With fine-grained search space introduced in Section \ref{sec:fine_search_space},
model compression can be view as a special case of neural architecture search.
With a unified representation of search space, neural architecture, hyperparameters, and transformers can be searched at the same time.

VEGA provides unified interfaces for components of deep learning frameworks.
It supports three different deep learning frameworks as back-ends: Pytorch, Tensorflow and
Mindspore.
Developers can choose their favorite back-end to implement their algorithms.
These algorithms can be training on different hardware platforms, such as CPU,
GPU and Ascend.
Besides, VEGA provides model conversion tools that can help the model to be
easily deployed on different hardware, such as GPUs, smart mobiles, and Ascend.
Then, it's easy to collect feedback of models on different hardware, for example, inference latency.

\subsection{Pipeline}
\label{sec:pipeline}
Building a deep learning system can be formulated as several steps of a pipeline.
For a complete pipeline, the input is the dataset and constraints of models (search space and search
algorihtm) and the output is the final model.
VEGA defines five basic steps and they are summarized as follows together with their corresponding supported algorithms:
\begin{enumerate}
\item \textbf{Neural architecture search (NAS)}: 
	
CARS~\cite{yang2020cars}. It utilizes continuous evolution method to efficiently search for a serial of architectures with only 0.4 GPU days. In each iteration of evolutionary algorithm, current samples can directly inherit the parameters from the supernet and parent samples, which effectively improves the efficiency of evolution. CARS can obtain a series of models with different sizes and precisions in one search.

SM-NAS~\cite{YaoLewei19}. It presents a two-stage coarse-to-fine searching strategy named Structural-to-Modular NAS (SM-NAS) for searching a GPU-friendly design of both an efficient combination of modules and better modular-level architecture for object detection. Specifically, structural-level searching stage first aims to find an efficient combination of different modules; Modular-level searching stage then evolves each specific module and pushes the Pareto front forward to a faster task-specific network.

SP-NAS~\cite{jiang2020sp}. It proposes a two-phase serial-to-parallel architecture search framework named SP-NAS towards a flexible task-oriented detection backbone. Specifically, the serial-searching round aims at finding a sequence of serial blocks with optimal scale and output channels in the feature hierarchy by a Swap-Expand-Reignite search algorithm; the parallel-searching phase then assembles several sub-architectures along with the previous searched backbone into a more powerful parallel-structured backbone.

Auto-Lane~\cite{xu2020curvelane}. It proposes a lane-sensitive architecture search framework named CurveLane-NAS to automatically capture both long-ranged coherent and accurate short-range curve information while unifying both architecture search and post-processing on curve lane predictions via point blending.

Adelaide-EA~\cite{nekrasov2019fast}. It focuses on searching for high-performance compact segmentation architectures, able to run in real-time using limited resources. To achieve that, It intentionally over-parameterises the architecture during the training time via a set of auxiliary cells that provide an intermediate supervisory signal and can be omitted during the evaluation phase.

SR-EA~\cite{zhang2019aim}. It reviews the AIM 2019 challenge on constrained example-based single image super-resolution with focus on proposed solutions and results. 

ESR-EA~\cite{song2020efficient}. It constructs a modular search space, takes the parameters and computations as constraints, and the network accuracy (PSNR) as the objective to search for a lightweight and fast super-resolution model. In this way, the network structure is hardware friendly by compressing the super-resolution network from three aspects: channel, convolution, and feature scale.

AutoFIS~\cite{liu2020autofis}. It proposes a two-stage algorithm called Automatic Feature Interaction Selection (AutoFIS). AutoFIS can automatically identify important feature interactions for factorization models with computational cost just equivalent to training the target model to convergence.

AutoGroup~\cite{liu2020autogroup}. It uses AutoML to seek useful high-order feature interactions to train on without manual feature selection. For this purpose, an end-to-end
model, AutoGroup, is proposed, which casts the selection of feature interactions as a structural optimization problem.
	
\item \textbf{Model Compression}:

QEA. Quantization based on Evolutionary Algorithm (QEA) is an automatic hybrid bit quantization algorithm. It uses an evolutionary strategy to search for the quantization bit width of each layer in a CNN network.

EA-Pruning~\cite{wang2018towards}. EA-Pruning applies evolutionary method to automatically eliminate redundant convolution filters by representing each compressed network as a binary individual of specific fitness. The population of compressed networks is upgraded at each evolutionary iteration using genetic operations. As a result, a Pareto Front of compact CNN will be given based on the fitness factor, such as flops and accuracy.

\item \textbf{Hyperparameter optimization (HPO)}:
	
ASHA~\cite{li2018system}. It introduces a simple and robust hyperparameter optimization algorithm called ASHA, which exploits parallelism and aggressive early-stopping to tackle large-scale hyperparameter optimization problems.

BOHB~\cite{falkner2018bohb}. It proposes to combine the benefits of both Bayesian optimization and bandit-based methods, in order to achieve the best of both worlds, and proposes a new practical state-of-the-art hyperparameter optimization method, which consistently outperforms both Bayesian optimization and Hyperband on a wide range of problem types

BOSS~\cite{huang2020asymptotically}. It proposes an efficient and robust bandit-based algorithm called Sub-Sampling (SS) in the scenario of hyperparameter search evaluation, and evaluates the potential of hyperparameters by the sub-samples of observations and is theoretically proved to be optimal under the criterion of cumulative regret.

\item \textbf{Data Augmentation}:
	
PBA~\cite{ho2019population}. It introduces a new data augmentation algorithm, Population Based Augmentation (PBA), which generates nonstationary augmentation policy schedules instead of a fixed augmentation policy. 

CycleSR~\cite{chen2020unsupervised}. It proposes a novel framework which is composed of two stages: 1) unsupervised image translation between real LR images and synthetic LR images; 2) supervised super-resolution from approximated real LR images to HR images. It takes the synthetic LR images as a bridge and creates an indirect supervised path from real LR images to HR images. Any existed deep learning based image super-resolution model can be integrated into the second stage of the proposed framework for further improvement.

\item \textbf{Fully Train}:
	
Disout~\cite{tang2020beyond}. Disout perturbs the neuron outputs instead of the direct set-to-zero operation used in Dropout [], which is more flexible and can effectively reduce the complexity of the model while minimizing the damage to the network representation, thereby improving the generalization of the network. It considers a flexible and universal neuron output disturbance by minimizing the Empirical Rademacher Complexity optimization. The results show the priority to the conventional Dropout.

Auto Augment~\cite{wei2020circumventing}. It delves deep into the working mechanism, and reveals that AutoAugment may remove part of discriminative information from the training image and so insisting on the ground-truth label is no longer the best option.
\end{enumerate}

\subsection{Unified Interfaces}
Mature deep learning frameworks have similar components, such as dataset
loader, training modules, model generator, trainer/evaluator.
There are different implementations for each component, but their
functionalities are the same.
So each component can be abstracted into a unified interface.
These interfaces are summarized as follows:
\begin{enumerate}
\item \textbf{Dataset}: Data are always loaded from different kinds of files on
	disks and converted into tensors for different back-ends.
	VEGA defines the \_\_getitem\_\_ interface to load an item of the dataset
	into NumPy format, like Pytorch.
	An adapter is used to convert the NumPy data into the corresponding tensor for
	different back-end.
	For a new dataset, we only need to register the new dataset class in ClassFactory.

\item \textbf{Training modules}: These modules are independent of network architectures and contain optimizer, lr\_scheduler, and loss functions.
	The interfaces of these modules are abstracted in the style of PyTorch interfaces.

\item \textbf{Model generator}: Different search space and algorithm can have
	different ways to generate neural architectures.
	The to\_model interface should be implemented in a network description class
	for a given search space and algorithm.
	A model description is framework independent and  has a uniformed format which
	contains the structure information of a model and construction parameters.
	The information is sampled from search space and can be used to generate
	an instance model.

\item \textbf{Training process}: VEGA supports the standard training process for
	three different frameworks and abstracts it into a trainer.
	A new training process can inherit the base trainer and override the
	interface train\_process.
	It's easy to introduce new steps or functionalities in the training process.
	
\end{enumerate}

\subsection{Configurability}
VEGA makes all components of AutoML algorithms can be configured by a
hierarchical structure.
All information independent of algorithms, e.g. resources can be configured in the `general' item.
The item of `pipeline' indicates the activated steps and the sequence of these steps.
The components of each step are configured under its corresponding name.
The components include step type, dataset, search
algorithm, search space, model description, and so on.
All the information can be extracted in a configuration file.
More details are introduced in Section \ref{sec:search_space}.

The following example shows the configuration of HPO algorithm AshaHpo \cite{li2018system} on Cifar10 \cite{krizhevsky2009learning}.

{\scriptsize{}\begin{lstlisting}
general:
    worker:
        devices_per_job: 1
pipeline: [hpo]
hpo:
    pipe_step:
        type: NasPipeStep
    dataset:
        type: Cifar10
    search_algorithm:
        type: AshaHpo
    search_space:
        type: SearchSpace
    trainer:
        type: Trainer
        epochs: 10
    model:
        model_desc:
            ...
\end{lstlisting}}{\scriptsize\par}

%
%
%

\subsection{Distributed Searching}
\begin{figure}
  \centering
  \includegraphics[scale=0.40]{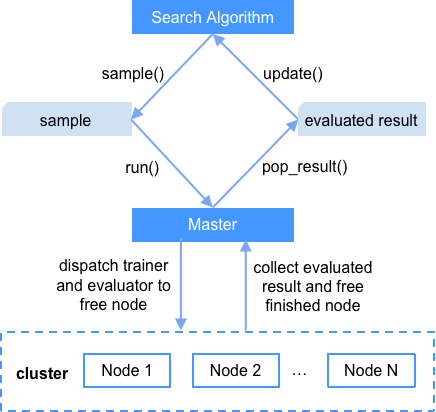}
    \vspace{-3mm}
  \caption{The distributed searching mechanism of VEGA. Search algorithm samples instances of the search space and submits training tasks to the master node. The master assigns these tasks to cluster nodes and collects the evaluated results. Then search algorithm gathers the feedback from the master. }
  \label{fig:distributed_search}
\end{figure}

VEGA abstracts the training process into a trainer and can adopt off-the-shelf distributed training mechanism of back-ends for training.
For searching, we design a distributed searching system that can easily manage the dispatch of tasks and the gather of evaluated results.
The mechanism is illustrated in Figure \ref{fig:distributed_search}.
`Search Algorithm' samples instances of search space and receives feedbacks from trained models.
The master node assigns training tasks to cluster nodes and collects the evaluated results.
Sampling and training are decoupled, then the searching process can be scaled to large clusters.



\section{Fine-grained Search Space}

\begin{figure*}
  \centering
  \includegraphics[scale=0.25]{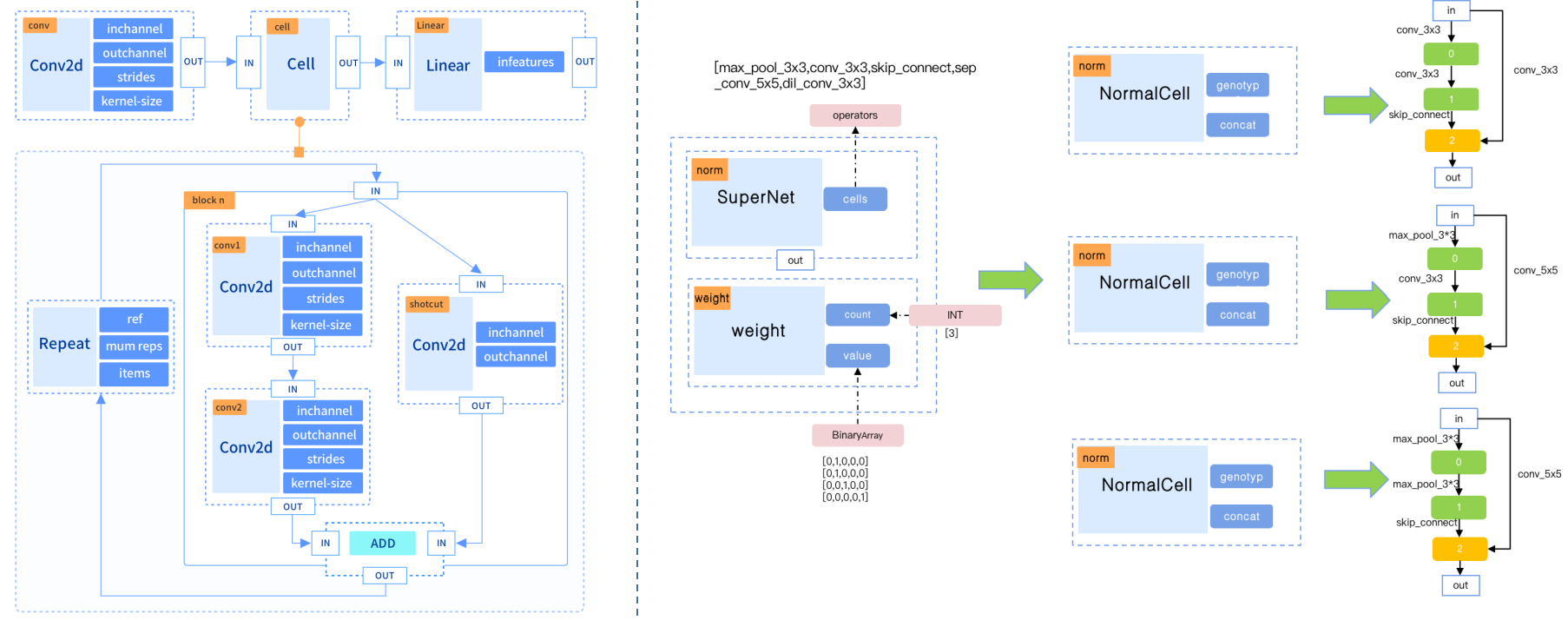}
  \caption{Overview of VEGA Fine Grained Search Space.
  A serial of wrappers for different operators and modules are defined and configurable variables are abstracted as the corresponding attributes.
  Left part shows how fine-grained
	modules are assembled into a network for sample based neural architecture
	search methods. Right part shows how to define a parameter-sharing
	supernet.}
  \label{fig:search_space}
\end{figure*}
\label{sec:search_space}
Different AutoML algorithms may have different search spaces.
These search spaces have great diversity and a lack of a common representation.
Data augmentation aims to find an optimal combination of tensor transformers and
get the magnitude for each tensor transformer.
These options can be viewed as hyperparameters.
The search space of data augmentation and HPO can be easily represented for that the key and range of value can be well defined as described in Section \ref{sec:coarse_search_space}.
The search space is independent of the concrete search algorithm.
However, the structure information in NAS has a more complex relationship than that in data augmentation or HPO.
VEGA introduces a fine-grained representation to decouple search space and search algorithms through a name binding mechanism.
More details will be introduced in Section \ref{sec:fine_search_space}.

\subsection{Search Space for HPO, Data Augmentation}
\label{sec:coarse_search_space}
For a well-defined search space, the search scope, range, and sampling mode should be determined.
VEGA defines ParamTypes for each item in the search scope, which has three attributes: key, type, and range.
`Key' indicates the variable to be searched.
`Type' means the type of variable.
`Range' shows the range of the variable. 
Sometimes, `Type'  also indicates the sampling mode.
For example, `INT' means the variable is an integer, and `INT\_CAT’ stands for
that values in the range are treated as different categories, an integer value is uniformly sampled.
`FLOAT\_EXP’ means the variable is float and exponentially sampled from a range.
Basic types defined in VEGA includes `INT', `INT\_EXP', `INT\_CAT', `FLOAT', `FLOAT\_EXP', `FLOAT\_CAT', and `STRING'.
The range of values can be a list and also be an interval depending on the sampling mode.

In general, items in the search scope can have complex dependencies. 
The values of an item can be dependent on that of another item.
VEGA defines some conditions to express the relationships between the values of different items during sampling.


\begin{itemize}
\item \textbf{Sampling}:  VEGA defines a standard sampling process, including three steps:
1. Cast: convert the configuration information into a ParamType.
2. Encode: encode the discrete values to continuous space.
Then, discrete and continuous values can be processed in a unified way.
3. Decode: decode the encoded continuous values to original discrete forms.


\item \textbf{Condition}: 
VEGA use a directed acyclic graph (DAG) to construct this relationship. 
The node of DAG is an item in the search scope,  and the edge indicates the dependency between two items.
The child nodes will be activated, only after the parent node meets defined conditions.
Currently, VEGA supports the following conditions: 
1. `EQUAL': When the sampled value of the parent node equals the configured value, the child nodes will be activated. 
2. `NOT\_EQUAL':  It's the opposite case of `EQUAL'.
3. `IN': When the sampled value of the parent node is in the configured range, the child nodes will be activated.
4. `FORBIDDEN' : When the parent node contains a defined value, the child nodes are disabled.

\end{itemize}

The following configuration shows a joint optimization for data augmentation and HPO:
{\scriptsize{}\begin{lstlisting}
search_space:
    type: SearchSpace
    hyperparameters:
    	key: dataset.batch_size
        type: INT_CAT
        range: [8, 16, 32, 64, 128, 256]
        key: dataset.tensformers
        type: STRING
        range: ['Cutout', 'Rotate','Brightness','Color']
        key: trainer.optim.params.lr
        type: FLOAT_EXP
        range: [0.00001, 0.1]
        key: trainer.optim.type
        type: STRING
        range: ['Adam', 'SGD']
        key: trainer.optim.params.momentum
        type: FLOAT
        range: [0.0, 0.99]
        key: network.custom.G1_nodes
        type: INT
        range: [3, 10]
        key: network.custom.G1_K
        type: INT
        range: [2, 5]
        key: network.custom.G1_P
        type: FLOAT
        range: [0.1, 1.0]
    condition:
        key: condition_for_sgd_momentum
        child: trainer.optim.params.momentum
        parent: trainer.optim.type
        type: EQUAL
        range: ["SGD"]
\end{lstlisting}}{\scriptsize\par}

\subsection{Search Space for NAS}
\label{sec:fine_search_space}

The search space for neural architecture search is complicated, every
hyperparameter or each module in the architecture can be in search scope, such as input and output channels for a convolution
operator, the number of convolution operators of a sub-module, connections between different modules.
Besides, the representation of operators may be relevant to the back-end framework.
VEGA defines a serial of wrappers for different operators and modules.
From coarse to fine, they are displayed as follows:

\begin{itemize}
\item \textbf{Network}: It is a wrapper for general network structure and can
	represent any neural architecture.
	Some basic attributes are defined in this wrapper, such as input channels,
	resolutions or the class number for classification tasks.
\item \textbf{Cell}: A cell is a component of a network and can be stem, body or head.
	It's used to assemble blocks.
	Block type and input information, connection, and number of repetitions are defined in cell wrapper.
\item \textbf{Block}: A block is composed of basic operators, for example, residual block.
	Operator types, input parameters, and connections can be specified.
\item \textbf{Operator}: It is a unified wrapper for basic operators for different frameworks. It provides the same way to get the hyperparameters and weights.
\end{itemize}

VEGA assigns a unique name for each module(network, cell, block, operator), and
each hyperparameter can be indexed by a name.
This mechanism decouples the search space and network definition in the actual backend.
Neural architecture can be configured through the given structure names with configuration files or python code.
We show that this fine grained search space can represent two different kinds of neural architecture search methods, sample-based NAS and parameter-sharing NAS.

\textbf{Sample-based NAS}: The sample-based algorithm is expected to modify architecture attributes by sample a value from search space.
First, we need to use the fine-grained modules in VEGA, as shown in Figure \ref{fig:search_space}, ResNet is assembled by fine-grained modules.
Second, we extend the type of the hyper parameters to cover most sampling scenarios.
The following is an example, searching the items of `doublechannel' and
`downsample inchannels' in variant block of ResNet \cite{YaoLewei19}.
{\scriptsize{}\begin{lstlisting}
model:
    modules ['resnet']
    resnet:
        type: ResNet
        
search_space:
    type: SearchSpace
    hyperparameters:
        -   key: resnet.cell.inchannels
            type: MutilyPositionArray
            range: [64]
            lenth: 8
            times: 3
            n: 2
        -   key: resnet.cell.strides
            type: IntArray
            range: (1,2)
            length: 8
                
\end{lstlisting}}{\scriptsize\par}

In this example, `MutilyPositionArray' stands for change the value at the specified position of the array.
Four parameters in `MutilyPositionArray', `range' indicates the initial value of this array, `length' indicates the array size.` times' indicates how many index positions are sampled from index list of this array.
`n' indicates in sampled positions, the value will be multiplied by n.
`IntArray range=[(0, 16), (0, 32)]' to construct an array with length 2.
The first value of the array samples an integer from 0 to 16, and the second value samples an integer from 0 to 32.
Finally, we obtain the hyperparameter by fine-grained name. `resnet.cell.inchannels' indicates the input parameter `inchannels' of the cell in the ResNet. 
For the cell-level module, additional attributes such as `convs'  provided to obtain information of all Conv2d operators in this cell, all the information will be recorded into a list.
A fine-grained ResNet network can be used for all search algorithms of Sample Base.

To prune channels of all Con2d operators in the ResNet, that is to change the values of `inchannels' and `outchannels' of Conv2d operators, we can define hyper parameters as below:

{\scriptsize{}\begin{lstlisting}
model:
    modules ['resnet']
    resnet:
        type: ResNet
        
search_space:
        type: SearchSpace
        hyperparameters:
            -   key: resnet.convs.inchannels
                type: IntArray
                range: [(0,16), (0,16), (0,16), (0,32), 
                    (0,32), (0,32), (0,32), (0,32)]
            -   key: resnet.convs.outchannels
                type: IntArray
                range:[(0,16), (0,16), (0,16), (0,32),
                    (0,32), (0,32), (0,32), (0,32)]
\end{lstlisting}}{\scriptsize\par}

\begin{table*}
      \vskip -2mm
    \caption{Our searched models for Ascend have a great advantage over RegNets, ResNets and EfficientNets. The batch size of inference on Ascend chips and GPU is 1 and 64, respectively. \textbf{The less inference time the better.}}  
    \centering
      \vskip 0.1in
      \begin{small}
        \begin{tabular}{l|ccc|ccc}
        \toprule
        Model & Resolution & Params(M) & Flops(B) & Top-1 & Time(Ascend)(ms)  & Time(V100)(ms) \\
        \midrule
        DNet-155  & 224 & 15.24 & 1.55 &  \bf 74.6\% &  \bf 2.53 &  \bf 14.54 \\  
		RegNetX-600MF \cite{radosavovic2020designing} & 224 & 6.2 & 0.6 &   74.1\% & 4.12 & 17.03  \\
        
		\midrule
         DNet-175  & 224 & 19.48 & 1.75 & \bf 75.7\% &  \bf 3.66 &  \bf 18.68 \\  
		RegNetX-800MF \cite{radosavovic2020designing} & 224 & 7.26 & 0.8 &   75.2\% & 4.13 & 22.10  \\
                        
        \midrule
         DNet-261  & 224 & 21.42 & 2.61 & \bf 77.3\% &  \bf 3.64 &  \bf 25.31 \\       
		RegNetX-1.6GF \cite{radosavovic2020designing} & 224 & 9.19 & 1.6 &   77.0\% & 5.54 & 34.98  \\
        
        \midrule
         DNet-424  & 224 & 25.31 & 4.24 &  \bf 78.5\% &  \bf 3.56 & 35.19 \\       
		ResNet-50 \cite{HeZRS16} & 224 & 25.56 & 4.09 &   76.1\% & 4.05 & 54.41  \\
		EfficientNet-B0 \cite{tan2019efficientnet} & 224 & 5.29 & 0.39 &   77.7\% & 12.35 &  \bf 29.83  \\

        \midrule
         DNet-501  & 224 & 28.47 & 5.01 &  \bf 79.2\% &  \bf 3.60 & \bf 39.57 \\       
		EfficientNet-B1 \cite{tan2019efficientnet} & 240 & 7.79 & 0.69 &   78.7\% & 12.75 & 48.93  \\
		ResNet-152 \cite{HeZRS16} & 224 & 60.19 & 11.51 &   78.3\% & 8.46 & 132.68  \\
         
        \midrule
         DNet-533  & 224 & 44 & 5.33 &  \bf 81.1\% &  \bf 5.82 &  \bf 63.61 \\       
		EfficientNet-B2 \cite{tan2019efficientnet} & 260 & 9.11 & 0.99 &   80.4\% & 23.29 & 63.87  \\
             
        \midrule
         DNet-1816  & 224 & 103.00 & 18.16 & \bf 82.4\% &  \bf 11.61 & 131.92 \\       
		EfficientNet-B3 \cite{tan2019efficientnet} & 300 & 12.23 & 1.83 &   81.5\% & 46.23 &  \bf 106.35  \\

        \midrule
         DNet-2346  & 224 & 130.45 & 23.46 &  82.8\% &  \bf 20.87 & \bf 191.89 \\       
		EfficientNet-B4 \cite{tan2019efficientnet} & 380 & 19.34 & 4.39 &   \bf 83.0\% & 69.23 & 223.13  \\
         
        \midrule
        DNet-4520  & 224 & 246.6 & 45.20 &  83.5\% &  \bf 19.27 &  \bf 275.91 \\
		EfficientNet-B5 \cite{tan2019efficientnet} & 456 & 30.39 & 10.27 &   \bf 83.6\% & 195.88 & 453.85  \\
        
        \midrule
        DNet-4520-320  & 320 & 246.6 & 92.25 & \bf 84.2\% &  31.663 &  \bf 506.16 \\
	 	EfficientNet-B6 \cite{tan2019efficientnet} & 528 & 43.04 & 19.07 & 84.0\% & - & 785.25  \\
                         
        \bottomrule
        \end{tabular}
        \end{small}
    \label{tab:imagenet}
          \vskip -2mm
\end{table*}

\textbf{Parameter-Sharing NAS}: The Parameter-Sharing algorithm defines selections of the connections and operators by training a weight. First, we use the fine-grained network SuperNet and specify genotype and concat. Second, we define the SearchSpace to obtain the weight of the SuperNet, the weight module has two attributes: `count' indicates the max count of weight array, that means how many nodes in one cell. `value' indicates the value of weight in each node, the type of value is `BinaryArray' indicates the value is a binary matrix and each row contains a unique value of 1, and the others are 0.

{\scriptsize{}\begin{lstlisting}
model:
    modules ['darts']
    darts:
        type: SupperNet
        cells:
        	genotype: ['none', 
        	  'max_pool_3x3', 
        	  'avg_pool_3x3', 
        	  'skip_connect', 
        	  'sep_conv_3x3', 
        	  'sep_conv_5x5', 
        	  'dil_conv_3x3', 
        	  'dil_conv_5x5']
        	concat: [2, 3, 4, 5]
        	
search_space:
        type: SearchSpace
        hyperparameters:
            -   key: darts.weight.count
                type: INT_CAT
                range: [3]
            -   key: darts.weight.value
                type: BinaryArray
                range: [4, 5]
\end{lstlisting}}{\scriptsize\par}

In this way, the search algorithm such as CARS \cite{yang2020cars} can directly change the weights of cells by key `darts.weight.value'. And select the connection with the best value of weights.

\begin{figure}
      \vskip -2mm
  \centering
  \includegraphics[scale=0.17]{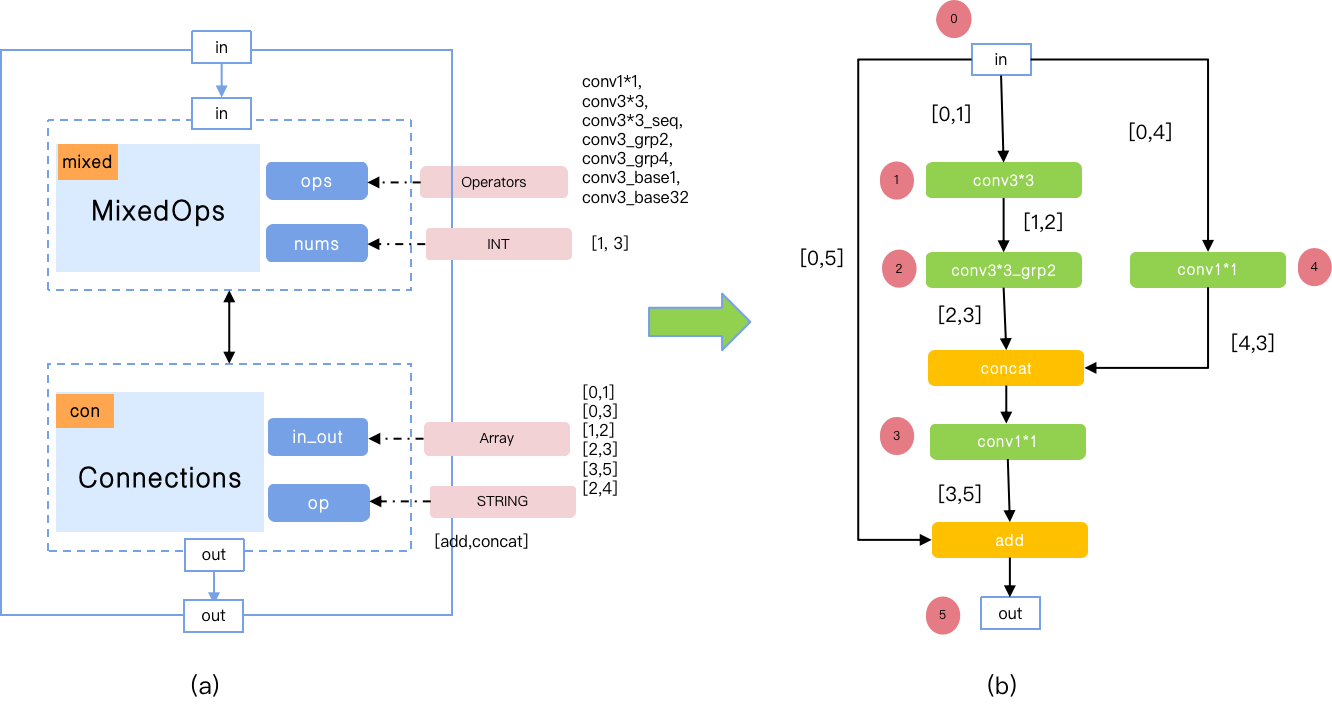}
    \vspace{-3mm}
  \caption{Abstract structure and an instance of DNet block.
	(a) There are 7 basic operators allowed in the stem of this structure. The number of these operators is no more than 3. Skip connections can be added between any disjunct operators. There is a default connection between the input and output of this block.
	(b) illustrates an instance with 3 operators in the stem. There is a skip connection between input and the third operator and an extra `conv1*1' operator is inserted to
	accommodate the resolutions of feature maps.}
  \label{fig:cell_structure}
        \vskip -2mm
\end{figure}

\begin{figure}
      \vskip -1mm
  \centering
  \includegraphics[scale=0.28]{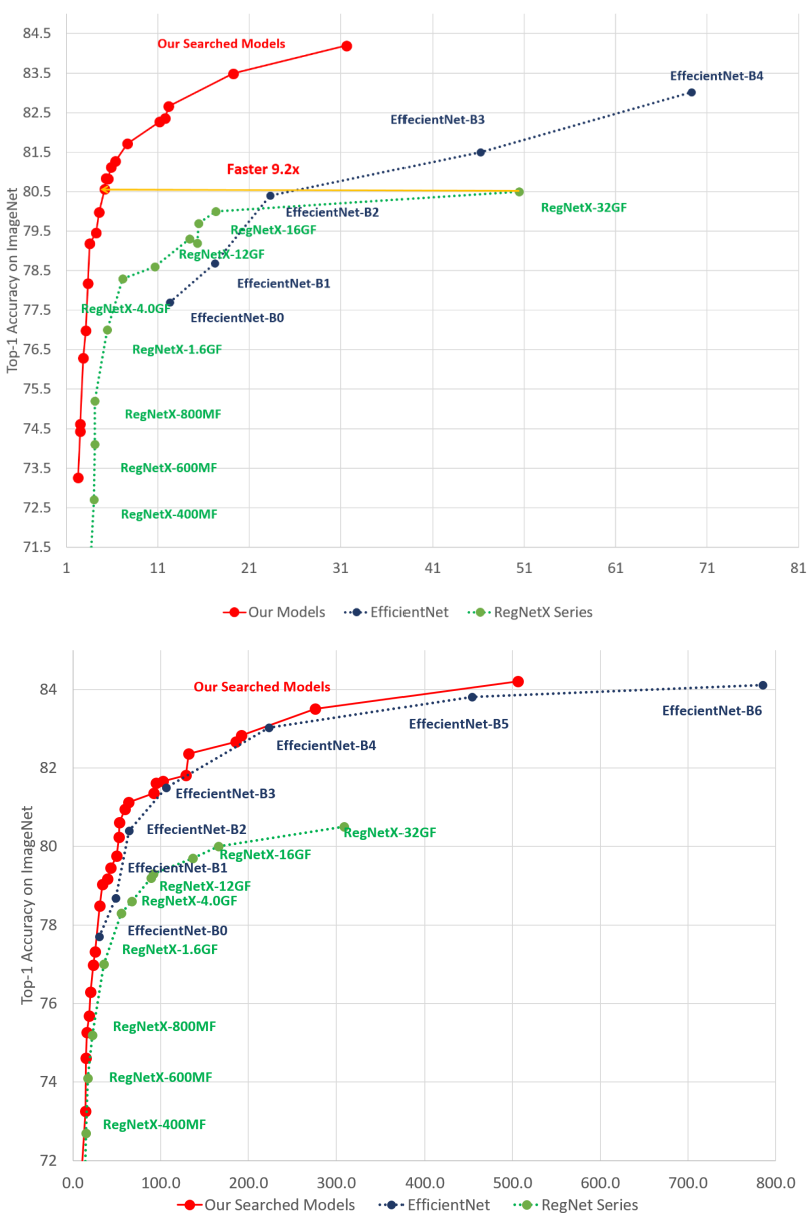}
  \vspace{-3mm}
  \caption{The model in searched DNet model zoo for Ascend can be $10\times$ faster then EfficientNet-B5, and $9.2\times$ faster than RegNetX-32GF on Ascend under same performance on ImageNet.
  These models are also efficient on GPU and even have a slight advantage over EfficientNets.}
  \label{fig:dnet_results}
        \vskip -1mm
\end{figure}

\subsection{Example: DNet Backbone Search Space}
Inspired by Residual blocks \cite{HeZRS16} and MobileNet blocks \cite{SandlerHZZC18},
we intend to find a series of hardware friendly blocks and construct efficient architectures on Ascend(Davinci Architecture), which is named `DNet'.
Figure \ref{fig:cell_structure}(a) illustrates the abstract block structure of DNet which is represented by a block wrapper in fine-grained search space.
There are 7 basic operators allowed in the stem and 5 options for the ratio of output channel size and input channel size.
This block wrapper has two key attributes: one is for placeholders of operators and the other is for connections between operators.
An instance of this structure as illustrated in Figure \ref{fig:cell_structure}(b) contains 
at most 3 operators in the stem of the block structure.
Skip connections can be added between any disjunct nodes, including input/output nodes.
Multiple input streams for an operator are combined by an element-wise `Add' or channel 
`Conat' operator into a single input as the final input.
Each operator is followed by a `BatchNorm' operator.
A `ReLU' operator is inserted after `Add' or `Concat' and `BatchNorm' that isn't followed by `Add' or `Concat'.
To facilitate the repetition of a block, we introduce a constraint that the output channel size is the same as the input channel size.

\section{Experiments}

\subsection{Application 1: Search for Hardware Efficient Backbones: DNet Model Zoo}
We attempt to construct a model zoo of Pareto optimized hardware-efficient backbones for downstream tasks with VEGA.
We adopt the two-level searching policy of the search algorithm SM-NAS \cite{YaoLewei19} for DNet and conduct all experiments on ImageNet \cite{deng2009imagenet}.
In micro-level searching, we consider more than 800k valid blocks.
We choose the first 100 classes, and for each class, we use 500 images as the training set and 100 images as the validation set.
With this small proxy dataset, architectures generated by effective blocks tend to have similar accuracies.
So we search on original ImageNet with 1000 classes in macro-level searching.
In all search processing, we evaluate each model the accuracy on ImageNet and inference time on Ascend and GPU(V100).

 Finally, we acquire a bunch of hardware friendly networks for both Ascend and GPU, as illustrated in Figure \ref{fig:dnet_results}. The model in our searched model zoo for Ascend can be $10\times$ faster then EfficientNet-B5 \cite{tan2019efficientnet}, and $9.2\times$ faster than RegNetX-32GF \cite{radosavovic2020designing} on ImageNet.
The searched models are also efficient on GPU and have a slight advantage over EfficientNets \cite{tan2019efficientnet}.
More details of some models compared to RegNets \cite{radosavovic2020designing} , ResNets \cite{HeZRS16} and EfficientNets are displayed in Table \ref{tab:imagenet}. 
It can be seen that our models are more efficient than these widely used baselines. The model zoo and the pretrained models have been released for the community.

 \begin{figure}
       \vskip -2mm
  \centering
  \includegraphics[scale=0.5]{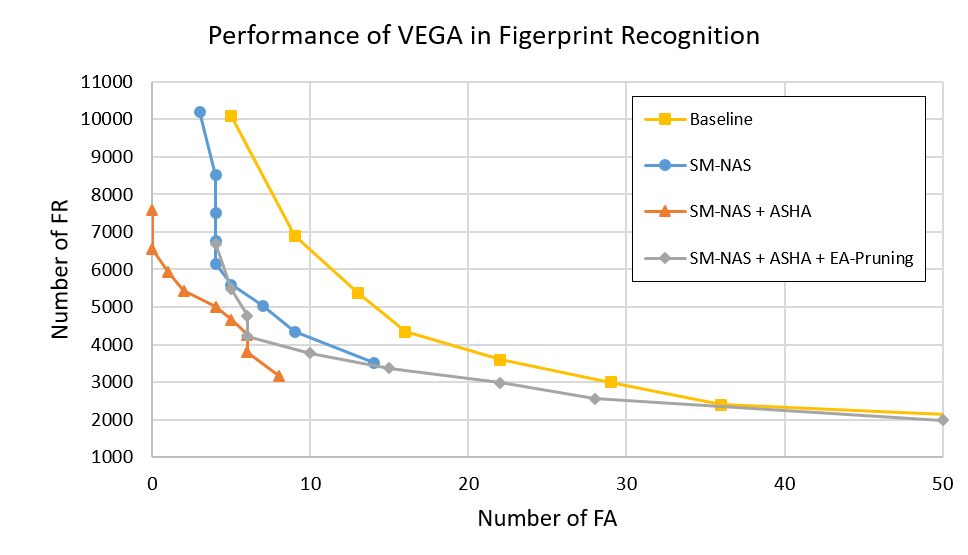}
    \vspace{-6mm}
  \caption{The comparison results of different combinations of AutoML algorithms applied into fingerprint recognition. Multiple components of AutoML algorithms can be optimized together and acquire better performance.
  The smaller value of false acceptance and false rejection the better. }
  \label{fig:fingerprint}
        \vskip -3mm
\end{figure}

\begin{table}
      \vskip -2mm
    \caption{We implement CARS \cite{yang2020cars} with VEGA and evaluate the models on  Cifar-10\cite{krizhevsky2009learning}.
    Three type models of different sizes are reported in the original paper.
    With the same test error, models searched by VEGA have smaller parameter size.
    These models achieve less test error.\textbf{The lower test error the better.}}  
    \centering
        \begin{small}
        \begin{tabular}{l|cc}
        \toprule
        Model   & Test Error  &  Params(M) \\
        \midrule
        CARS-1\cite{yang2020cars} &  2.95\% & 2.9 \\
        CARS-1(VEGA) & 2.95\% & \bf 2.2  \\
        
         \midrule
        CARS-2\cite{yang2020cars} &  2.86 \% & 3.0 \\
        CARS-2(VEGA)&  \bf 2.75\% & 2.4  \\
        
         \midrule
        CARS-3\cite{yang2020cars} &  2.79\% & 3.1 \\
        CARS-3(VEGA)& \bf 2.7\% & 3.7  \\
        
        \bottomrule
        \end{tabular}
        \end{small}
    \label{tab:cars}
          \vskip -2mm
\end{table}

\begin{table}
	\renewcommand\arraystretch{1.1}
	\vskip -2mm
	\caption{Classification results using Disout~\cite{tang2020beyond} on CIFAR-10, CIFAR-100 and ImageNet datasets.}  
	\centering
	\begin{small}
		\begin{tabular}{c|cc}
			\hline
			   & CIFAR-10 accuracy  &  CIFAR-100 accuracy \\
			\hline
			ResNet-56 &  94.50\% & 73.71\% \\
			\hline
			 & ImageNet Top-1 accuracy & Top-5 accuracy\\
			\hline
			ResNet-50 &  78.76\% & 94.33\% \\
			
			\hline
		\end{tabular}
	\end{small}
	\label{tab:disout}
	\vskip -2mm
\end{table}

\begin{table}
	\renewcommand\arraystretch{1.}
	\vskip -2mm
	\caption{Classification results using QEA on CIFAR-10 dataset.}  
	\centering
	\begin{small}
		\begin{tabular}{c|cc}
			\toprule
			ResNet-20 & accuracy  &  BOPs \\
			\midrule
			QEA-1 &  90.46\% & 189.2M \\
			\midrule
			QEA-2 &  90.65\% & 255.2M \\
			\midrule
			QEA-3 &  91.26\% & 396.8M \\
			
			\bottomrule
		\end{tabular}
	\end{small}
	\label{tab:qea}
	\vskip -2mm
\end{table}

\begin{table}
	\renewcommand\arraystretch{1.}
	\vskip -2mm
	\caption{Classification results using EA-Pruning~\cite{wang2018towards} on ImageNet dataset.}  
	\centering
	\begin{small}
		\begin{tabular}{c|cc}
			\toprule
			& AlexNet & VGGNet \\
			\midrule
			Compression Ratio &  39x & 46x \\
			\midrule
			Speed-up Ratio &  25x & 9.4x \\
			\midrule
			Top-1 Accuracy &  58.4\% & 70.3\% \\
			\midrule
			Top-5 Accuracy & 80.8\%& 89.6\%\\
			\bottomrule
		\end{tabular}
	\end{small}
	\label{tab:eapruning}
	\vskip -2mm
\end{table}

\begin{table}
      \vskip -2mm
    \caption{SP-NAS \cite{jiang2020sp} can be used for pedestrian detection. 
    We evaluate this algorithm on dataset ECP \cite{braun2019eurocity} 
    with the metric log average miss-rate (LAMR).
    Our implemented algorithm can get lower LAMR in all cases, 
    including reasonable, small, occluded and all.
    \textbf{The lower LAMR the better.}}  
    \centering
        \begin{small}
        \begin{tabular}{l|cccc}
        \toprule
        Method & reasonable & small & occluded & all \\
        SPNet(Original) & 0.054 & 0.110 & 0.252 & 0.165 \\
        \midrule
        SPNet(Vega) & \bf 0.042 & \bf 0.095 & \bf 0.216 & \bf 0.139 \\
        \bottomrule
        \end{tabular}
        \end{small}
    \label{tab:ecp}
          \vskip -2mm
\end{table}

\begin{table}
      \vskip -2mm
    \caption{AutoLane \cite{xu2020curvelane} aims to find models with high F1 score evaluated on CULane \cite{pan2018SCNN}.
    The flops of our searched models are larger than that reported in original paper, but they get higher F1 scores. }  
    \centering
        \begin{small}
        \begin{tabular}{l|cc}
        \toprule
        Method & Flops (G) & F1 Score (\%) \\
        \midrule
        CULane-M\cite{xu2020curvelane} & \bf 35.7 &  73.5 \\
        \midrule
        CULane-M(VEGA) & 66.9 & \bf 74.6 \\
        \midrule
        CULane-L\cite{xu2020curvelane} & \bf 86.5 & 74.8 \\
        \midrule
        CULane-L(VEGA) & 273 & \bf 75.2 \\
        \bottomrule
        \end{tabular}
        \end{small}
    \label{tab:culane}
      \vskip -2mm
\end{table}

\begin{table}
      \vskip -2mm
    \caption{We implement two recommended search algorithms auto-group \cite{liu2020autogroup} and auto-fis \cite{liu2020autofis} on VEGA.
     Compared with FM \cite{rendle2010factorization} and DeepFM \cite{guo2017deepfm}, auto-group and auto-fis have higher AUC(Area under curve) metric on dataset Avazu.
    \textbf{The higher AUC the better.} }  
    \centering
        \begin{tabular}{l|cc}
        \toprule
        Method & AUC \\
        
        \midrule
        FM \cite{rendle2010factorization} & 0.7793\\

        \midrule
        DeepFM \cite{guo2017deepfm} & 0.7836\\
        \midrule
        Auto-group \cite{liu2020autogroup} & \bf 0.7900 \\
        \midrule
        Auto-group(VEGA) & \bf 0.7900 \\
        \midrule
        Auto-FIS\cite{liu2020autofis} & 0.7852 \\
        \midrule
        Auto-FIS(VEGA) & \bf 0.7880 \\

        \bottomrule
        \end{tabular}
    \label{tab:avazu}
          \vskip -2mm
\end{table}

\begin{table}
      \vskip -2mm
    \caption{ESR-EA \cite{song2020efficient} and SR-EA \cite{zhang2019aim} have been implemented in VEGA for super-resolution.
    The searched models are trained with dataset DIV2K \cite{Timofte_2017_CVPR_Workshops} and  and evaluated with Urban 100.
    \textbf{The higher PSNR the better.}}  
    \centering
        \begin{small}
        \begin{tabular}{l|ccc}
        \toprule
        Method & Model Size (M) & Flops (G) & PSNR \\
        \midrule
        ESRN-V (article) & \bf 0.32 & 73.4 & 31.79 \\
        \midrule
        ESRN-V1 & 1.32 & 40.616 & 31.65 \\
        \midrule
         ESRN-V2 & 1.31 & 40.21 &  37.84 \\
        \midrule
        ESRN-V3 & 1.31 &  41.676 &  37.79 \\
        \midrule
         ESRN-V4 & 1.35 &  40.17 & \bf 37.83 \\
        \midrule
        SRx2-A & 3.20 &  196.27 & \bf 38.06 \\
        \midrule
         SRx2-B & 0.61 &  35.03 &  37.73 \\
        \midrule
         SRx2-C & \bf 0.24 & \bf 13.29 &  37.56 \\
        \bottomrule
        \end{tabular}
        \end{small}
    \label{tab:urban100}
      \vskip -2mm
\end{table}

\subsection{Application 2: Fingerprint Recognition in Smart Phone}
To achieve better performance with a smaller model for fingerprint recognition task in the smartphone, we applied our VEGA pipeline to optimizing the whole pipeline for both NAS and HPO. 
We attempt to apply the algorithms listed in Section ~\ref{sec:pipeline} to the fingerprint recognition task.
In this task, the fingerprint samples are classified into two categories based on the user's characteristics.
The objective is to get a better balance between the number of false acceptance (FA) and the number of false rejection (FR).

We conduct NAS with SM-NAS \cite{YaoLewei19}, HPO with ASHA \cite{li2018system} and model compression with EA-Pruning \cite{wang2018towards} in these tasks.
These are four comparisons of experiments: (1) baseline, (2) with SM-NAS only, (3) with SM-NAS and ASHA and (4) SM-NAS, ASHA and EA-Pruning.
The baseline is a manual designed model and default hyperparameters.
We optimize the learning rate and learning rate decay for HPO.

Figure ~\ref{fig:fingerprint} shows the comparisons between different pipe steps of VEGA. It can be found that the composition of SM-NAS and ASHA outperforms others.
SM-NAS significantly improves the model performance compared with the baseline.
With HPO, the model searched by SM-NAS achieves much better results with smaller FA and FR. Thus using the whole pipeline of VEGA can help different tasks.

\subsection{VEGA Benchmarks}
VEGA has a great range of application scenarios.
We list some benchmarks implemented in VEGA for different tasks. We show that our VEGA is capable of optimizing models for different tasks including image classification, pedestrian/lane detection, click-through rate prediction, and image super-resolution.
The output models and results are comparable or better than those reported in the original paper.

\textbf{Image Classification.} We implement CARS \cite{yang2020cars} with VEGA and evaluate the models on  Cifar-10\cite{krizhevsky2009learning}. It can be found that the searched models from VEGA perform better than the originally reported results in terms of accuracy and number of parameters. We also report our VEGA’ s results of Disout~\cite{tang2020beyond}, QEA and EA-Pruning~\cite{wang2018towards} on the CIFAR-10, CIFAR-100 and ImageNet datasets. The results show that these algorithms perform well on our pipeline.

\textbf{Pedestrian Detection.} Table \ref{tab:ecp} shows our results on VEGA comparing to the original reported results in \cite{jiang2020sp}. 
Note that SPNAS method won 1st place in the ECP dataset \cite{braun2019eurocity} which is one of the largest pedestrian detection datasets.

\textbf{Lane Detection.} We report our VEGA's results for the lane detection task on the CULane dataset \cite{pan2018SCNN} in Table \ref{tab:culane}. The Autolane \cite{xu2020curvelane} method reaches the SOTA performance on CULane dataset while we performs even better. CULane dataset \cite{pan2018SCNN} is one of the biggest Lane detection dataset.

\textbf{Click Through Click Prediction.} Click through rate prediction is an important task in a modern recommendation system which aims at predicting the customer's click behavior and rank the items for better recommendation. We examined our VEGA on the famous Avazu \footnote{\url{https://www.kaggle.com/c/avazu-ctr-prediction/}} dataset in Table \ref{tab:avazu}. It can be found that our pipeline can perform well.

\textbf{Image Super-Resolution.} Table \ref{tab:urban100} reports the performance of VEGA pipeline on the image super-resolution task. We report the Peak signal-to-noise ratio (PSNR) on Urban100 dataset \cite{huang2015single}. It can be found that the searched models perform even better than the competition models in terms of Flops and PSNR.

\section{Conclusion}
We have presented an efficient and comprehensive AutoML framework,  VEGA.
It integrates various key modules of AutoML with more than 20 algorithms, including neural architecture search (NAS), hyperparameter optimization (HPO), data augmentation, model compression, and fully train. 
With our novel design of search space, the NAS search space can be configured readily and the NAS search algorithms can be carried out seamlessly.
Besides, VEGA provides a unified front-end interface for training frameworks and supports different deep learning frameworks. VEGA can further be trained and deployed on different hardware platforms. Experiments on various benchmarks and tasks show the performance superiority of VEGA. We are continuously improving VEGA towards an efficient, scalable, configurable AutoML Pipeline.




\bibliography{ref}
\bibliographystyle{mlsys2020}


\end{document}